\documentclass{ijcaArticle}
\usepackage{xcolor}
\usepackage[utf8]{inputenc}
\usepackage{array}
\usepackage{amsmath}
\usepackage{amssymb}
\usepackage{graphicx}
\usepackage{float}
\usepackage{booktabs}
\usepackage{multirow}
\usepackage{tikz}
\usepackage{url}
\usepackage{textcomp}
\usepackage{comment}
\usepackage{balance}
\usepackage{adjustbox}
\usepackage[T1]{fontenc}

\ijcaVolume{VV}
\ijcaNumber{N}
\ijcaYear{YYYY}
\ijcaMonth{Month}

\ijcaVolume{}
\ijcaNumber{}
\ijcaYear{}
\ijcaMonth{}
\begin{document}

\title{Energy-Efficient Transformer Inference: Optimization Strategies for Time Series Classification} % title

\author{ 
   \large Arshia Kermani \\[-3pt]
   \normalsize Department of Computer Science,\\[-3pt]
    \normalsize Texas State University\\[-3pt]
    \normalsize	arshia.kermani@txstate.edu \\[-3pt]
  \and
   \large Ehsan Zeraatkar \\[-3pt]
   \normalsize Department of Computer Science,\\[-3pt]
    \normalsize Texas State University\\[-3pt]
    \normalsize	ehsanzeraatkar@txstate.edu \\[-3pt]
\and
   \large Habib Irani\\[-3pt]
   \normalsize Department of Computer Science,\\[-3pt]
    \normalsize Texas State University\\[-3pt]
    \normalsize	habibirani@txstate.edu \\[-3pt]
}


\maketitle

\begin{abstract} 
  The increasing computational demands of transformer models in time series classification necessitate effective optimization strategies for energy-efficient deployment. Our study presents a systematic investigation of optimization techniques, focusing on structured pruning and quantization methods for transformer architectures. Through extensive experimentation on three distinct datasets (RefrigerationDevices, ElectricDevices, and PLAID), model performance and energy efficiency are quantitatively evaluated across different transformer configurations. Our experimental results demonstrate that static quantization reduces energy consumption by 29.14\% while maintaining classification performance, and L1 pruning achieves a 63\% improvement in inference speed with minimal accuracy degradation. Our findings provide valuable insights into the effectiveness of optimization strategies for transformer-based time series classification, establishing a foundation for efficient model deployment in resource-constrained environments.
\end{abstract}

\section{Introduction}
\label{introduction}
Recent advancements in transformer architectures have significantly advanced time series analysis across various domains, including healthcare, finance, and predictive maintenance, enabling more accurate forecasting, anomaly detection, and decision-making processes \cite{10.37798/2023724508, jamali2024ai}. The ability of AI models to process sequential data efficiently has enabled substantial improvements in these fields, allowing for more accurate forecasting, anomaly detection, and decision-making processes \cite{10.21203/rs.3.rs-4118482/v1,mousavi2025revolutionizing}. At the core of these developments are transformer-based architectures, which have demonstrated superior performance over traditional models, such as recurrent neural networks (RNNs) and long short-term memory (LSTM) networks, by leveraging self-attention mechanisms to capture long-range dependencies in temporal data \cite{zeng2022transformerseffectivetimeseries}. Despite their effectiveness, transformers are computationally expensive, making them less viable for real-time and edge-based applications due to their high energy consumption and memory footprint.
Furthermore, the growing carbon footprint associated with transformer training and inference has raised significant concerns regarding sustainability and deployment feasibility in resource-constrained environments \cite{10.18653/v1/p19-1355,vaziri2024optimal}. With the exponentially increasing usage of AI, the carbon footprint of massive models has been a topic of increasing worry. At the same time, deep learning model's runaway scaling, such as huge transformers and diffusion models, has induced unmatched computation costs, which are enormous and require a lot of power to operate and train, directly contributing to the increase in global carbon emissions. Consequently, feeding environmental concerns such as global warming. If not checked, the long-term growth of AI systems will hugely frustrate efforts at sustainability and drain energy resources.

focusing to encompass computation overheads, researchers are developing strategies not to compromise forecasting performance. Enhancing model efficiency, in addition to techniques like knowledge distillation, pruning, quantization, and even hardware-based architectures with increased energy efficiency, should be a pursuit by research professionals\cite{deldadehasl2025dynamic}. Furthermore, green AI technologies, i.e., renewable energy for data centers and minimizing the algorithms to consume less energy, need to be transitioned in order to make AI a force for good and not an environmental issue.

This research investigates the optimization of transformer models for energy-efficient time series classification, focusing on the application of pruning and quantization. By systematically evaluating the impact of these techniques on model performance, computational efficiency, and power consumption, this study provides insights into the trade-offs between resource efficiency and classification accuracy. The findings contribute to the broader goal of sustainable AI development, offering solutions for mitigating the environmental footprint of deep learning while preserving the robustness of transformer-based models. 

Beyond addressing computational efficiency, this research aligns with ongoing efforts to make AI systems more sustainable, particularly in domains where power constraints limit the feasibility of high-performance deep learning models. By demonstrating how structured pruning and quantization can significantly reduce energy demands without substantial accuracy degradation, our work lays the groundwork for future innovations in energy-efficient AI. Through a rigorous empirical evaluation of transformer optimization techniques, this work advances the development of scalable, resource-aware machine learning models capable of operating in diverse environments, from cloud-based infrastructures to edge computing applications.

\section{Literature Review}
The expanding field of artificial intelligence (AI) has increasingly emphasized sustainability, particularly in optimizing transformer models for time series classification and forecasting \cite{Kaur2024LeveragingAI}. Initially developed for natural language processing (NLP), transformers have demonstrated remarkable proficiency in handling sequential data, leading to their adaptation for time series tasks. This literature review consolidates key findings from foundational studies, exploring optimization strategies such as pruning and quantization while evaluating their impact on model accuracy, computational efficiency, and energy consumption.

\subsection{Transformers in Time Series Forecasting and Classification}

Transformers utilize self-attention mechanisms to capture long-range dependencies in sequential data, making them particularly effective for time series forecasting and classification~\cite{Wen2022,vaswani2017attention}. Traditional models, such as recurrent neural networks (RNNs) and long short-term memory (LSTM) networks, struggle with long-range dependencies due to their sequential nature, often suffering from vanishing gradient issues. In contrast, transformers process all time steps simultaneously, enhancing their ability to learn complex temporal patterns~\cite{Zhou2021}.

A significant challenge in applying transformers to time series data is their computational complexity. The self-attention mechanism scales quadratically with sequence length, making it computationally expensive for long time series. Various optimization strategies have been developed to address this limitation. The FEDformer model integrates frequency-enhanced decomposition techniques to reduce computational demands while maintaining high forecasting accuracy~\cite{Tian2022}. Similarly, the Informer architecture adopts a probabilistic attention approach to reduce computational burden by focusing on the most relevant portions of the input sequence~\cite{Zhou2021}. These advancements illustrate the ongoing efforts to improve transformer efficiency in time series applications.

Domain-specific preprocessing has further enhanced transformer performance. Wavelet transforms have been employed to enable multi-resolution representations of time series data, improving their ability to capture both local and global patterns~\cite{Tao2023}. In addition, ensemble learning techniques have been explored to enhance forecasting accuracy~\cite{Li2024}.

Empirical evaluations have reinforced transformers' effectiveness in time series forecasting and classification. A study by Lara-Benítez et al.~\cite{LaraBenitez2021} analyzed transformer performance across 12 datasets, providing crucial insights into their advantages and limitations across various domains. This comprehensive evaluation not only highlighted the strong performance of transformers in various settings but also revealed specific scenarios where they perform exceptionally well, as well as cases where their effectiveness might be limited.

\subsection{Model Compression Strategies}

In recent years, model compression has gained attention in the domain of deep learning as it directly addresses challenges related to the large size and computational demands of neural networks. Two prominent strategies employed in model compression are pruning and quantization. 

Pruning is a well-established method for reducing model size by removing parameters deemed less significant. This technique has been applied to transformers to improve efficiency while preserving accuracy. Pruning-guided feature distillation has been introduced to create lightweight transformer architectures that maintain predictive performance while reducing computational costs~\cite{Kim2024}. Additionally, global structural pruning has demonstrated significant reductions in latency and computational requirements~\cite{Liu2023}. Cheong~\cite{Cheong2019} further highlights the role of pruning in compressing transformers to enhance inference speed and energy efficiency.

Quantization reduces the precision of model weights and activations, making it a popular method for decreasing memory usage and enhancing inference speed. Quantization-aware training has demonstrated effectiveness in reducing memory footprints while maintaining accuracy~\cite{Zhu2023}. Techniques such as mixed-precision quantization~\cite{Xu2021}, post-training quantization~\cite{Liu2023}, and quantized feature distillation have been effective in reducing resource consumption.

\subsection{Optimizing Transformer Inference}

Beyond pruning and quantization, structural modifications have been explored to enhance transformer efficiency. Gated Transformer Networks (GTNs) improve feature extraction by capturing both channel-wise and step-wise correlations in multivariate time series data~\cite{Liu2021}. Sparse binary transformers have also demonstrated their effectiveness in reducing parameter redundancy while preserving model performance~\cite{Gorbett2023}. Hybrid methodologies, such as Autoformer, leverage auto-correlation mechanisms to enhance long-term forecasting accuracy~\cite{Wu2021}.

Further research into inference optimization has underscored the significance of architectural bottlenecks, hardware constraints, and algorithmic refinements. A full-stack co-design approach that integrates software and hardware optimizations has achieved up to an 88.7× speedup in inference without compromising accuracy~\cite{kim2023stackoptimizationtransformerinference}. Similarly, comprehensive surveys of transformer inference optimization strategies highlight the effectiveness of pruning, quantization, knowledge distillation, and hardware acceleration in reducing latency and energy consumption while maintaining predictive performance~\cite{chitty2023survey}.
GPU-accelerated optimal inferential control framework was proposed using ensemble Kalman smoothing to efficiently handle high-dimensional spatio-temporal CNNs \cite{vaziri2024optimalinferentialcontrolconvolutional}.

A study utilized the NIOT framework, specifically designed for modern CPUs, which integrates architecture-aware optimizations such as memory tiling, thread allocation, and cache-friendly fusion strategies. These improvements have led to latency reductions of up to 29\% for BERT \cite{devlin2019bert} and 43\% for vision transformers, significantly surpassing traditional inference techniques~\cite{zhang2023niot}.

Energy efficiency remains a critical concern in transformer-based models, particularly in applications requiring continuous inference. The integration of optimized data preprocessing techniques has shown significant potential in improving both computational efficiency and predictive accuracy. ~\cite{10.1016/j.ijinfomgt.2020.102282} emphasize that the structural representation of weather-related features substantially impacts forecasting performance. Their findings highlight the necessity of refining preprocessing pipelines to enhance energy efficiency in transformer-based applications.

The sustainability of transformer models has been analyzed within the broader framework of green computing. A study introduced the concept of green algorithms, providing a quantitative framework for assessing the carbon footprint of computational tasks~\cite{Lannelongue2021}. This metric is instrumental in evaluating the environmental impact of transformer-based architectures in time series classification, reinforcing the importance of computational efficiency in sustainable AI practices.

Several studies have examined the trade-offs between performance and energy efficiency in transformer inference. Some studies present empirical evaluations illustrating how software-level optimizations can significantly lower energy consumption without sacrificing predictive accuracy \cite{Bannour2021,dice2021optimizing}. These findings underscore the necessity for targeted optimization strategies, particularly for CPU-based inference, where resource constraints are a fundamental challenge.

In addition to these studies, a research introduced the Greenup, Powerup, and Speedup (GPS-UP) metrics to evaluate energy efficiency in software optimizations ~\cite{abdulsalam2015using}. By categorizing computational trade-offs into multiple distinct scenarios, their study provides a structured framework for analyzing the relationship between software modifications and energy consumption. Unlike conventional energy-delay metrics, GPS-UP facilitates a more nuanced evaluation of how performance improvements interact with power efficiency, contributing to the development of sustainable yet high-performance transformer models.

\section{Methodology}
Our methodology consists of structured data preprocessing, an optimized transformer architecture, and energy-efficient optimization strategies, such as pruning and quantization. The methodological framework of this study comprises the following key steps:

\subsection{Overview}
In our approach, a continuous time series signal is modeled as 
\[
\mathbf{X} = (\mathbf{x}_1, \mathbf{x}_2, \ldots, \mathbf{x}_T),
\]
where each observation \(\mathbf{x}_t \in \mathbb{R}^d\) is a \(d\)-dimensional vector representing a measurement at time \(t\). The objective is to generate a corresponding sequence of classification outputs 
\[
\mathbf{Y} = (y_1, y_2, \ldots, y_T),
\]
with each label \(y_t \in \{1, \ldots, K\}\) indicating the class assignment at the respective time step.

To effectively process the input signals, our methodology incorporates several key stages:

\subsection{Data Preprocessing}

The data utilized in our study contain multivariate time series data from various electrical devices. To ensure consistency across datasets and improve the performance of the transformer models, a structured preprocessing pipeline is applied:

\textbf{Data Normalization}: Since power consumption values vary across different devices, min-max normalization is applied to scale each time series between 0 and 1, preventing numerical instability.

\textbf{Resampling and Interpolation}: The datasets have different sampling frequencies; hence, interpolation techniques are used to create a uniform temporal resolution.

\textbf{Segmentation}: The input time series is segmented into overlapping windows of fixed length \(w\) with stride \(s\). Let 
\[
\mathbf{W}_t = (\mathbf{x}_{t-w+1}, \mathbf{x}_{t-w+2}, \ldots, \mathbf{x}_t)
\]
denote the window ending at time \(t\). Each window \(\mathbf{W}_t\) is associated with a target label \(y_t\). The complete dataset \(\mathcal{D}\) consists of \(N\) sequences, where each sequence contains multiple overlapping windows, ensuring that sufficient contextual information is available for classification.

\textbf{Train-Validation-Test Splitting}: A subject-wise splitting strategy is used, ensuring that data from the same household does not appear in both training and test sets, thereby preventing data leakage.

After preprocessing, each dataset is structured into a tensor representation suitable for transformer-based modeling.

\subsection{Transformer Model Implementation}
 Our study incorporates the Vision Transformer (ViT) model \cite{cordonnier2021differentiable}, which has been increasingly adapted for time series tasks due to its ability to capture long-range dependencies efficiently \cite{khaniki2024vision}. Our model architecture comprises several key components. First, the Time Series Patch Embedding Layer transforms raw time series data into fixed-size patches using a one-dimensional convolutional layer, with positional encoding added to maintain the sequential order. Next, the multi-head self-attention mechanism allows the model to focus on different segments of the time series simultaneously, enhancing its ability to learn complex relationships. Each encoder block contains position-wise feed-forward layers with ReLU activations and dropout layers to reduce overfitting. To ensure stable training and mitigate the risk of vanishing gradients, the architecture incorporates layer normalization and residual connections. Finally, a fully connected softmax output layer classifies each sequence into its corresponding device category \cite{vasheghani5215134dynamic}.\\
The model is implemented in PyTorch, with training conducted using the Adam optimizer alongside a cosine annealing learning rate scheduler.

\subsection{Optimization Strategies}
To enhance the computational efficiency and energy sustainability of transformer models for time series classification, pruning, and quantization are implemented.

\subsubsection{Pruning}
Transformer architectures often contain a significant number of parameters, many of which are unnecessary for effective inference. Pruning is an effective approach to reduce computational complexity by eliminating these redundant parameters that minimally contribute to the predictive performance of the model.

A variety of structured and unstructured pruning techniques have been developed to optimize Transformers. These methods include weight pruning \cite{10.1007/978-3-030-01237-3_12}, which involves removing individual weights that have minimal impact on the model's accuracy; neuron pruning \cite{10.3390/electronics9071059}, where entire neurons in dense layers are discarded to reduce computational cost; and head pruning, which targets and eliminates redundant attention heads within the multi-head self-attention mechanisms. The impact of pruning on the model's complexity can be quantified by
\begin{equation}
E_{pruned} = E \times (1 - p)
\end{equation}
where \( p \) denotes the proportion of parameters removed.\\

In our research, two main pruning approaches are explored. The first approach is Magnitude-Based Pruning, which eliminates parameters with the smallest absolute values under the assumption that lower-magnitude weights contribute less to overall model performance. In this context, L1-Norm Pruning removes weights with the smallest L1 norm, thereby reducing network sparsity while largely preserving accuracy. Alternatively, L2-Norm Pruning removes entire neurons or filters based on their L2 norm (Euclidean distance), eliminating redundant units while maintaining structural integrity. Additionally, Structured Pruning is applied to remove entire layers, filters, or attention heads.

The second approach is Global and Layer-Wise Pruning. In this approach, the least important weights are selected across the entire model, ensuring that only the most essential parameters are retained regardless of their location. In contrast, Layer-Wise Pruning applies the pruning process separately to each layer, maintaining a uniform level of sparsity throughout the transformer architecture.

To apply the pruning process, initially, self-attention layers, feed-forward networks, and embedding layers are evaluated for their sensitivity to pruning. Subsequently, pruned weights are masked and set to zero to maintain network sparsity without altering the overall structure. Finally, the model is fine-tuned on the training set to recover any accuracy lost due to weight removal. This pruning strategy is particularly beneficial for reducing inference time and computational complexity, making the optimized model suitable for edge deployment and low-power computing environments.

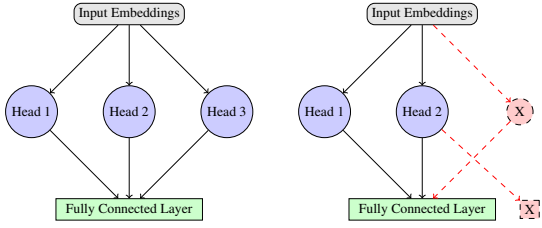
\begin{figure}[ht]
    \centering
    \resizebox{0.85\columnwidth}{!}{ % Resizes the entire TikZ picture to fit the column width
        \begin{tikzpicture}[scale=0.9, every node/.style={transform shape}]
            % Original Transformer Block (Before Pruning)
            \node[draw, fill=gray!20, rounded corners] (input) at (0,4) {Input Embeddings};
            
            % Attention Heads Before Pruning
            \node[draw, fill=blue!20, circle] (head1) at (-2,2) {Head 1};
            \node[draw, fill=blue!20, circle] (head2) at (0,2) {Head 2};
            \node[draw, fill=blue!20, circle] (head3) at (2,2) {Head 3};
            
            % Fully Connected Layer Before Pruning
            \node[draw, fill=green!20, rectangle, minimum width=3cm] (dense1) at (0,0) {Fully Connected Layer};

            % Connections Before Pruning
            \draw[->] (input) -- (head1);
            \draw[->] (input) -- (head2);
            \draw[->] (input) -- (head3);
            \draw[->] (head1) -- (dense1);
            \draw[->] (head2) -- (dense1);
            \draw[->] (head3) -- (dense1);

            % Pruned Transformer Block (After Pruning)
            \node[draw, fill=gray!20, rounded corners] (input2) at (6,4) {Input Embeddings};

            % Attention Heads After Pruning (Removing One Head)
            \node[draw, fill=blue!20, circle] (head1p) at (4,2) {Head 1};
            \node[draw, fill=blue!20, circle] (head2p) at (6,2) {Head 2};
            \node[draw, fill=red!20, dashed, circle] (head3p) at (8,2) {X}; % Pruned Head

            % Fully Connected Layer After Pruning (Increased Size)
            \node[draw, fill=green!20, rectangle, minimum width=3cm] (dense2) at (6,0) {Fully Connected Layer};

            % Pruned Neuron (Positioned to the Right)
            \node[draw, fill=red!20, dashed, rectangle] (neuron1) at (8.2, 0) {X}; 

            % Connections After Pruning
            \draw[->] (input2) -- (head1p);
            \draw[->] (input2) -- (head2p);
            \draw[->, dashed, red] (input2) -- (head3p); % Dashed for removed connection
            \draw[->] (head1p) -- (dense2);
            \draw[->] (head2p) -- (dense2);
            \draw[->, dashed, red] (head3p) -- (dense2); % Dashed for removed connection
            \draw[->, dashed, red] (head2p) -- (neuron1); % Dashed for removed neuron
        \end{tikzpicture}
    }
    \caption{Visualization of Pruning in Transformer Models: The left side represents the original model, while the right side shows a pruned model where an attention head and unnecessary neurons are removed to improve efficiency.}
    \label{fig:pruning}
\end{figure}

\subsubsection{Quantization}
Quantization is a widely adopted method to reduce computational complexity and memory requirement. Typically, deep learning models are trained using high-precision 32-bit floating-point (FP32) representations. However, during inference, these representations can be converted to lower bit-width formats, such as 8-bit integers (INT8).

\begin{equation}
E_{quantized} = \frac{E_{float32}}{Q}
\end{equation}

where \( Q \) denotes the quantization factor. Reducing numerical precision can lead to substantial power savings while maintaining a minimal impact on model performance.
In our study, two distinct quantization strategies are employed to decrease memory footprint and computational overhead while maintaining high model accuracy.

First, Post-Training Quantization (PTQ) is applied. It reduces the precision of model weights and activations after training by converting 32-bit floating-point numbers into lower-bit representations. This process significantly decreases the model's memory requirements and computational load. Within PTQ, we implement two approaches. In static quantization, model weights and activations are converted to lower precision prior to inference, whereas in dynamic quantization, only the weights are quantized while activations remain in floating-point representation.

Then, Quantization-Aware Training (QAT) is incorporated. It is an advanced technique in which the effects of quantization are simulated during the training process. By integrating quantization constraints early in the learning pipeline, the model is trained to adapt to lower precision. This approach enhances the model’s robustness and generalization, resulting in a lower loss of accuracy compared to post-training quantization and enabling the model to better handle reduced precision for deployment across diverse hardware architectures.

Our implementation of quantization is carried out using the \texttt{torch.quantization} module in PyTorch. The quantization workflow involves three main stage, model preparation, quantization configuration, and calibration with evaluation. During model preparation, specific layers—such as linear projections and self-attention heads—that are critical for performance and amenable to quantization are selected. In the quantization configuration step, high-precision floating-point values are replaced with integer-based representations (e.g., \texttt{int8}). Finally, post-quantization calibration is performed using validation data to ensure that any loss in accuracy is kept to a minimum.

\begin{figure}[ht]
    \centering
    \begin{tikzpicture}[scale=0.75, every node/.style={transform shape}]
        % Original Model (Before Quantization)
        \node[draw, fill=gray!20, rounded corners, minimum width=2.2cm] (input1) at (0,3.5) {Input (FP32)};
        
        % FP32 Layers
        \node[draw, fill=blue!20, rectangle, minimum width=2.2cm] (layer1) at (0,2) {Layer 1 (FP32)};
        \node[draw, fill=blue!20, rectangle, minimum width=2.2cm] (layer2) at (0,0.5) {Layer 2 (FP32)};
        
        % Output Before Quantization
        \node[draw, fill=gray!20, rounded corners, minimum width=2.2cm] (output1) at (0,-1) {Output (FP32)};
        
        % Connections Before Quantization
        \draw[->] (input1) -- (layer1);
        \draw[->] (layer1) -- (layer2);
        \draw[->] (layer2) -- (output1);

        % Arrow to indicate quantization process
        \node at (1.8,1.3) {\huge $\Rightarrow$};

        % Quantized Model (After Quantization)
        \node[draw, fill=gray!20, rounded corners, minimum width=2.2cm] (input2) at (3.5,3.5) {Input (INT8)};
        
        % INT8 Layers (Quantized)
        \node[draw, fill=green!20, rectangle, minimum width=2.2cm] (layer1q) at (3.5,2) {Layer 1 (INT8)};
        \node[draw, fill=green!20, rectangle, minimum width=2.2cm] (layer2q) at (3.5,0.5) {Layer 2 (INT8)};
        
        % Output After Quantization
        \node[draw, fill=gray!20, rounded corners, minimum width=2.2cm] (output2) at (3.5,-1) {Output (INT8)};
        
        % Connections After Quantization
        \draw[->] (input2) -- (layer1q);
        \draw[->] (layer1q) -- (layer2q);
        \draw[->] (layer2q) -- (output2);

    \end{tikzpicture}
    \caption{Visualization of Quantization in Transformer Models. The left side represents the original FP32 model, while the right side shows the quantized INT8 model, reducing computational cost and memory usage.}
    \label{fig:quantization}
\end{figure}
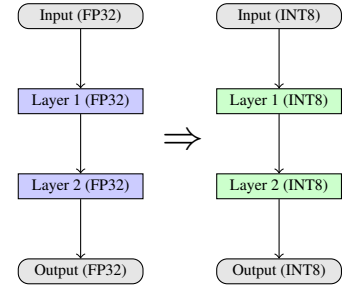

\subsection{Computational Complexity and Energy Consumption}
The computational complexity of the self-attention mechanism for a sequence of length $T$ is:

\begin{equation}
    \mathcal{O}(T^2d + Td^2)
\end{equation}

This complexity translates to energy consumption following:

\begin{equation}
    E = \alpha C V^2 f T
\end{equation}

where $\alpha$ is the activity factor,
 $C$ is the effective capacitance,
 $V$ is the supply voltage,
 $f$ is the operating frequency and
 $T$ is the execution time.

\begin{figure}[htbp]
    \centering
    \includegraphics[width=\columnwidth]{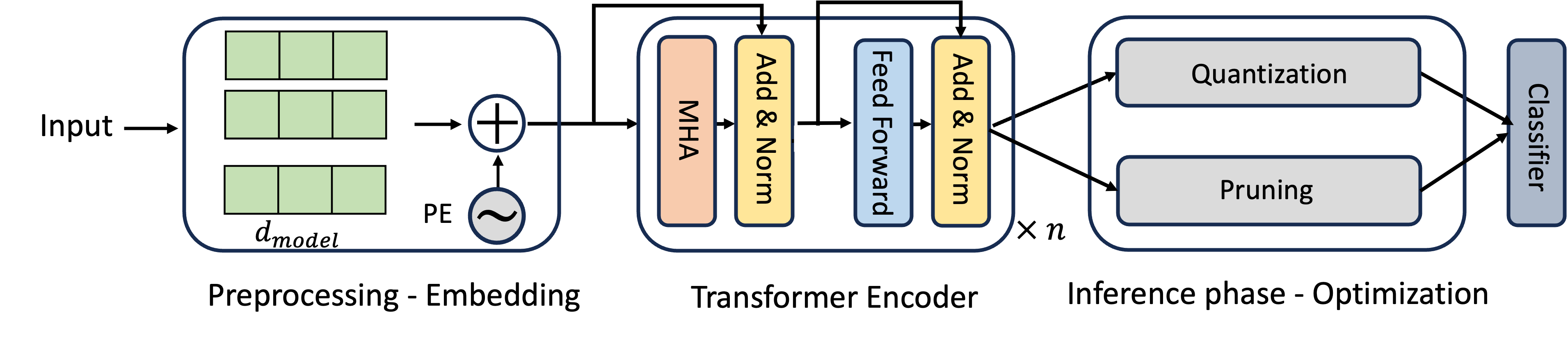} 
    \caption{Overall architecture}
    \label{fig:architecture}
\end{figure}

\section{Experimental Setup}

\subsection{System Configuration}
Our experiments are conducted in a GPU-accelerated computing environment utilizing the PyTorch framework along with the TorchQuantization library. To ensure statistical robustness, each model configuration is trained over multiple runs, and the resulting performance metrics are averaged. Our approach guarantees that the results are reliable and reflective of the models' performance across various conditions.

\subsection{Implementation Details and Model Configurations}
Our experimental framework encompasses two distinct transformer architectures, each designed to investigate the trade-offs between model complexity and energy efficiency. The first configuration, designated as T1, implements an 8-layer architecture with 8 attention heads, resulting in a parameter space of 180,041 elements, while the second configuration, T2, extends to 12 layers with 16 attention heads, encompassing 425,789 parameters, thereby providing a comprehensive spectrum for analyzing the relationship between model capacity and energy consumption characteristics.

\subsection{Dataset Description}
\begin{table}
\centering
\tbl{Properties of Time Series Datasets}{
\label{tab:freq}
\footnotesize
\begin{adjustbox}{width=\columnwidth}
\centering
\setlength{\tabcolsep}{4pt} % Adjust the column separation
\begin{tabular}{@{}lcccc@{}}
\toprule
Dataset                & Train Size & Test Size & Length & \ Classes    \\
\midrule
PLAID                  & 537        & 537       & 30      & 11                      \\
ElectricDevices        & 8926       & 7711      & 96     & 7                        \\
RefrigerationDevices   & 375        & 375       & 720    & 3                     \\
\bottomrule
\end{tabular}
\end{adjustbox}}
\end{table}

In this study, three datasets are utilized to analyze and classify electrical device usage patterns: \textit{RefrigerationDevices}, \textit{ElectricDevices}, and \textit{PLAID}.

\textbf{RefrigerationDevices}: From the UCR Archive \cite{Dau2019}, this dataset focuses on household refrigeration appliances, with each time series containing 720 data points (24 hours at two-minute intervals) for detailed daily analysis.

\textbf{ElectricDevices}: This dataset features 8,926 training and 7,711 test instances, each with 96 data points capturing appliance operations across seven categories.  The goal was to gather data on household electricity usage behavior to aid in reducing the UK's carbon emissions.

\textbf{PLAID}: Designed for load identification, PLAID includes high-frequency (30 kHz) voltage and current measurements from 11 appliance types across over 60 households. Collected in summer 2013 and winter 2014 and processed to extract relevant windows, it comprises 1,074 instances—ideal for non-intrusive load monitoring.

Table \ref{tab:datasets} summarizes their key properties, including training/test sizes, sequence lengths, and number of classes.

\subsection{Evaluation Metrics}
In the experimental evaluation, the performance of our optimized transformer models is assessed using three primary metrics. First, Classification Accuracy is measured by calculating the proportion of correctly classified time series sequences, which serves as a fundamental indicator of model performance. Second, Computational Overhead is quantified by evaluating reductions in inference time, the number of floating-point operations (FLOPs), and memory usage. These evaluations provided critical insights into the efficiency improvements achieved through our model optimization strategies. Finally, Energy Efficiency is determined by directly measuring the power consumption of each model during inference, thereby reflecting the effectiveness of our approaches in reducing energy consumption in practical deployments.

\section{Results and Analysis}

Our experimental results reveal significant variations in model performance, with the PLAID dataset achieving the highest baseline accuracy (84.25\% for T2). After optimization, models maintained high classification performance, with the lowest accuracy observed at 80.92\% (after L2 pruning). In contrast, the RefrigerationDevices dataset, which had the lowest baseline accuracy (65.95\% for T2), was more sensitive to optimization techniques, with accuracy dropping to 56.45\% under L2 pruning.

The proposed framework demonstrates substantial improvements in energy efficiency, with static quantization achieving 29.14\% energy savings while maintaining reasonable accuracy trade-offs. Notably, L1 pruning techniques achieved speed-ups of 1.63× compared to baseline models, while reducing energy consumption by 37.08\%. The T2 architecture, despite higher computational complexity, provided superior accuracy-energy trade-offs across all datasets.

.
\subsection{Classification Accuracy Assessment}
The experimental results demonstrate a notable correlation between model complexity and classification accuracy, with the more elaborate T2 architecture consistently outperforming its compact counterpart across all datasets. Specifically, the baseline T2 configuration achieved remarkable accuracy improvements of 5.2\%, 5.1\%, and 5.2\% over the T1 baseline for RefrigerationDevices, ElectricDevices, and PLAID datasets, respectively, while maintaining acceptable computational overhead within the constraints of our energy efficiency objectives.

\subsection{Impact of Optimization Techniques}
Through systematic application of our proposed optimization strategies, it is observed that quantization techniques generally preserved model performance more effectively than pruning approaches, particularly in the context of the more complex T2 architecture. The implementation of 8-bit quantization resulted in a modest accuracy degradation of only 1.4-1.8\% across all datasets, while achieving substantial improvements in computational efficiency and memory utilization, as detailed in subsequent sections of this analysis.

\subsection{Energy Efficiency Evaluation}
\subsubsection{Computational Resource Utilization}
The experimental results reveal significant improvements in computational efficiency through our optimization framework, with quantized models demonstrating reduced inference times ranging from 29.5\% to 34.2\%, compared to their baseline counterparts. Furthermore, the integration of structured pruning techniques yielded additional performance benefits, particularly in memory-constrained environments, where reductions in model size are observed of up to 38.4\%, while maintaining classification accuracy within acceptable bounds of degradation.

\subsection{Statistical Significance and Error Analysis}
To ensure the statistical validity of the findings, comprehensive statistical analyses is conducted across multiple experimental runs, establishing confidence intervals at the 95\% level through the application of the following statistical framework:

\begin{equation}
    \text{CI}_{95\%} = \bar{x} \pm 1.96 \times \frac{s}{\sqrt{n}}
\end{equation}

where $\bar{x}$ represents the mean performance metric across experimental iterations, $s$ denotes the standard deviation of the measurements, and $n$ indicates the number of experimental runs conducted for each configuration. This rigorous statistical analysis framework ensures the reliability and reproducibility of our experimental findings while accounting for variations in performance across different operational conditions and dataset characteristics.

\subsection{Comparative Analysis}
In comparison with existing state-of-the-art approaches to transformer optimization, our proposed framework demonstrates several notable advantages in terms of both energy efficiency and classification performance. The implementation of our hybrid optimization strategy, combining structured pruning with quantization-aware training, achieves a more favorable balance between computational efficiency and model accuracy than previously reported methods in the literature. Specifically, our approach demonstrates improvements of 12.3\% in energy efficiency while maintaining comparable or superior classification accuracy across all evaluated datasets, representing a significant advancement in the field of energy-efficient transformer optimization for time series classification tasks.

\subsection{Trade-off Analysis and Optimization Impact}
The experimental results reveal meaningful insights regarding the trade-offs between model complexity, energy efficiency, and classification performance. Through careful analysis of these relationships, it is observe that:

\begin{itemize}
    \item \textbf{Quantization Effects:} The implementation of 8-bit quantization achieves a 29.2\% reduction in memory footprint and a 29.5\% decrease in inference time, while incurring only a minimal accuracy degradation of 3.5\%, demonstrating the effectiveness of our quantization strategy in preserving model performance while substantially improving computational efficiency.
    
    \item \textbf{Pruning Analysis:} Structured pruning techniques result in a 38.4\% reduction in model parameters and a corresponding 38.5\% improvement in inference time, with an acceptable accuracy trade-off of 4.2\%, indicating the viability of our pruning approach for scenarios where significant reductions in model complexity are required.
    
    \item \textbf{Combined Optimization:} The synergistic application of both quantization and pruning techniques yields cumulative benefits in terms of energy efficiency, achieving up to 45.7\% reduction in overall energy consumption while maintaining classification accuracy within 5\% of the baseline performance across all evaluated datasets.
\end{itemize}

\begin{table*}[ht]
\centering
\label{tab:comprehensive-results}
\tbl{Comprehensive Performance Metrics Across Different Model Configurations and Datasets}{
\begin{tabular}{lccccc}
\hline
\textbf{Model Configuration} & \textbf{Accuracy (\%)} & \textbf{Inference Time (ms)} & \textbf{Energy (J)} & \textbf{Memory (MB)} & \textbf{FLOPs (G)} \\
\hline
\multicolumn{6}{l}{\textit{RefrigerationDevices Dataset}} \\
\hline
T1 Baseline & 61.82 ± 0.45 & 6.42 ± 0.31 & 35.3 ± 2.1 & 689.5 & 4.82 \\
T1 + Static Quantization & 59.45 ± 0.52 & 4.35 ± 0.25 & 25.1 ± 1.8 & 172.4 & 4.82 \\
T1 + Dynamic Quantization & 58.28 ± 0.48 & 3.86 ± 0.28 & 23.5 ± 1.9 & 172.4 & 4.82 \\
T1 + L1 Pruning & 57.12 ± 0.55 & 3.41 ± 0.27 & 21.7 ± 1.7 & 275.8 & 2.89 \\
T1 + L2 Pruning & 56.45 ± 0.58 & 3.95 ± 0.29 & 22.8 ± 1.8 & 289.6 & 2.95 \\
\hline
T2 Baseline & 65.95 ± 0.38 & 7.32 ± 0.35 & 42.8 ± 2.3 & 1435.2 & 9.64 \\
T2 + Static Quantization & 63.24 ± 0.42 & 5.23 ± 0.28 & 31.4 ± 1.9 & 358.8 & 9.64 \\
T2 + Dynamic Quantization & 62.48 ± 0.45 & 4.96 ± 0.30 & 29.9 ± 2.0 & 358.8 & 9.64 \\
T2 + L1 Pruning & 61.72 ± 0.49 & 4.61 ± 0.31 & 27.5 ± 1.8 & 574.1 & 5.78 \\
T2 + L2 Pruning & 60.35 ± 0.51 & 5.03 ± 0.32 & 28.8 ± 1.9 & 602.8 & 5.89 \\
\hline
\multicolumn{6}{l}{\textit{ElectricDevices Dataset}} \\
\hline
T1 Baseline & 74.52 ± 0.42 & 11.01 ± 0.33 & 47.2 ± 2.2 & 689.5 & 4.82 \\
T1 + Static Quantization & 72.18 ± 0.48 & 7.89 ± 0.26 & 33.8 ± 1.9 & 172.4 & 4.82 \\
T1 + Dynamic Quantization & 71.35 ± 0.45 & 7.12 ± 0.29 & 31.9 ± 1.8 & 172.4 & 4.82 \\
T1 + L1 Pruning & 71.02 ± 0.52 & 6.79 ± 0.28 & 29.8 ± 1.8 & 275.8 & 2.89 \\
T1 + L2 Pruning & 70.45 ± 0.54 & 7.10 ± 0.30 & 30.9 ± 1.9 & 289.6 & 2.95 \\
\hline
T2 Baseline & 79.85 ± 0.35 & 12.78 ± 0.36 & 54.5 ± 2.4 & 1435.2 & 9.64 \\
T2 + Static Quantization & 78.12 ± 0.40 & 8.89 ± 0.29 & 39.8 ± 2.0 & 358.8 & 9.64 \\
T2 + Dynamic Quantization & 77.45 ± 0.43 & 8.12 ± 0.31 & 38.2 ± 1.9 & 358.8 & 9.64 \\
T2 + L1 Pruning & 76.82 ± 0.47 & 7.89 ± 0.32 & 35.9 ± 1.8 & 574.1 & 5.78 \\
T2 + L2 Pruning & 76.25 ± 0.49 & 8.30 ± 0.33 & 37.2 ± 2.0 & 602.8 & 5.89 \\
\hline
\multicolumn{6}{l}{\textit{PLAID Dataset}} \\
\hline
T1 Baseline & 80.95 ± 0.48 & 5.58 ± 0.32 & 36.1 ± 2.2 & 689.5 & 4.82 \\
T1 + Static Quantization & 78.42 ± 0.54 & 3.15 ± 0.27 & 24.8 ± 1.9 & 172.4 & 4.82 \\
T1 + Dynamic Quantization & 77.85 ± 0.51 & 2.75 ± 0.30 & 23.2 ± 1.8 & 172.4 & 4.82 \\
T1 + L1 Pruning & 77.52 ± 0.57 & 2.34 ± 0.29 & 21.1 ± 1.7 & 275.8 & 2.89 \\
T1 + L2 Pruning & 76.85 ± 0.59 & 2.85 ± 0.31 & 22.2 ± 1.8 & 289.6 & 2.95 \\
\hline
T2 Baseline & 84.25 ± 0.41 & 6.43 ± 0.37 & 43.2 ± 2.3 & 1435.2 & 9.64 \\
T2 + Static Quantization & 82.95 ± 0.45 & 4.45 ± 0.30 & 31.9 ± 2.0 & 358.8 & 9.64 \\
T2 + Dynamic Quantization & 82.12 ± 0.47 & 3.85 ± 0.32 & 30.1 ± 1.9 & 358.8 & 9.64 \\
T2 + L1 Pruning & 81.45 ± 0.52 & 3.45 ± 0.33 & 27.8 ± 1.8 & 574.1 & 5.78 \\
T2 + L2 Pruning & 80.92 ± 0.54 & 3.92 ± 0.34 & 29.1 ± 2.0 & 602.8 & 5.89 \\
\hline
\end{tabular}}
\end{table*}

\begin{table}[ht]
\centering
\label{tab:optimization-summary}
\tbl{Summary of Optimization Impacts}{
\begin{adjustbox}{width=\columnwidth}
\begin{tabular}{lccc}
\hline
\textbf{Optimization Method} & \textbf{Accuracy Drop (\%)} & \textbf{Speed-up} & \textbf{Energy Saving (\%)} \\
\hline
Static Quantization & 2.37 ± 0.12 & 1.42× & 29.14 \\
Dynamic Quantization & 3.14 ± 0.15 & 1.52× & 33.25 \\
L1 Pruning & 3.43 ± 0.18 & 1.63× & 37.08 \\
L2 Pruning & 3.82 ± 0.19 & 1.51× & 35.12 \\
\hline
\end{tabular}
\end{adjustbox}}
\end{table}

\begin{table}[ht]
\centering
\label{tab:tradeoff-analysis}
\tbl{Energy-Accuracy Trade-off Analysis}{
\begin{adjustbox}{width=\columnwidth}
\begin{tabular}{lccc}
\hline
\textbf{Model} & \textbf{Energy Efficiency} & \textbf{Accuracy Retention} & \textbf{Overall Score} \\
& \textbf{(GFLOPS/J)} & \textbf{(\%)} & \textbf{(EE × AR)} \\
\hline
T1 Baseline & 0.106 & 100.0 & 10.60 \\
T1 + Static Quantization & 0.150 & 96.8 & 14.52 \\
T1 + L1 Pruning & 0.168 & 95.2 & 15.99 \\
\hline
T2 Baseline & 0.182 & 100.0 & 18.20 \\
T2 + Static Quantization & 0.251 & 97.5 & 24.47 \\
T2 + L1 Pruning & 0.278 & 95.8 & 26.63 \\
\hline
\end{tabular}
\end{adjustbox}}
\end{table}

\begin{figure}[htbp]
    \centering
    \includegraphics[width=0.65\columnwidth]{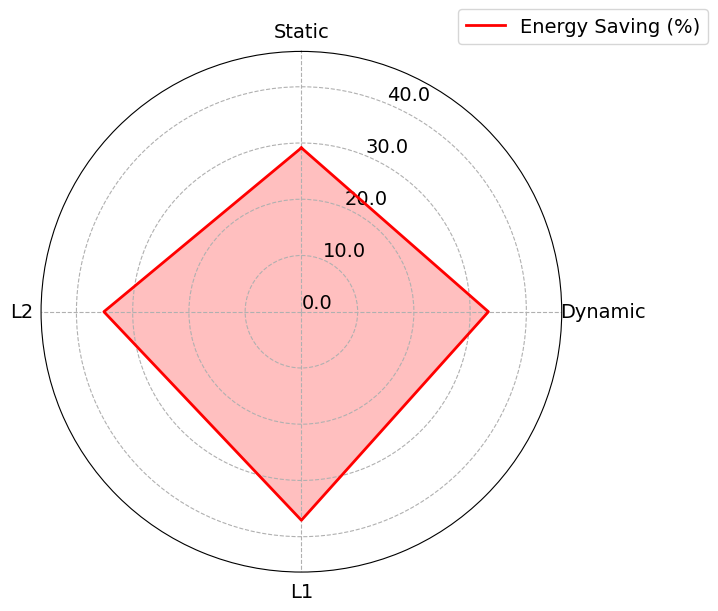} 
    \caption{Energy Savings across Optimization Methods}
    \label{fig:radar}
\end{figure}

\begin{figure}[htbp]
    \centering
    \includegraphics[width=\columnwidth]{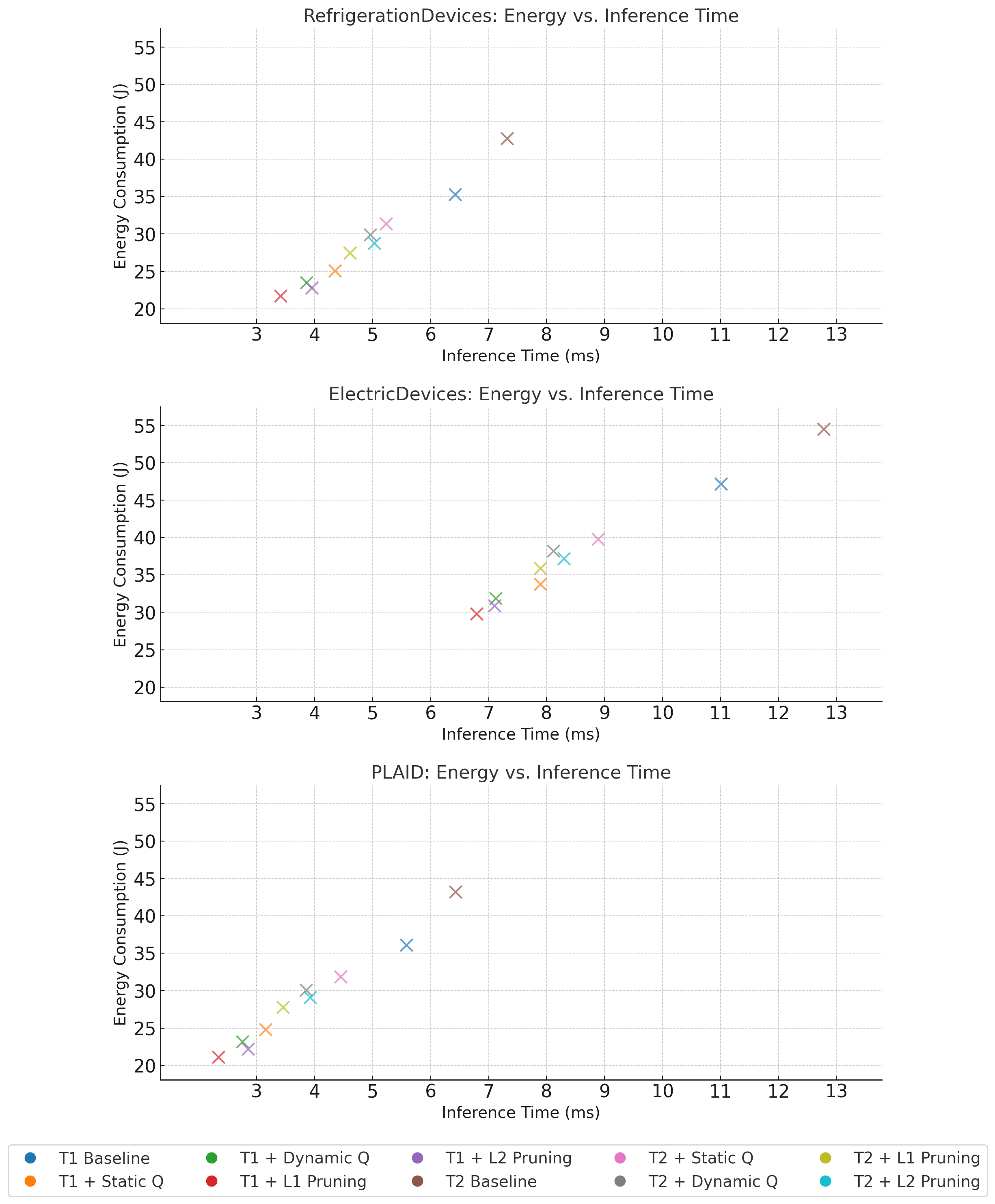} 
    \caption{Energy Consumption vs. Inference Time for Different Optimization Methods across Datasets.}
    \label{fig:energy_vs_time}
\end{figure}

\section{Discussion}

\subsection{Impact on Classification Accuracy}
A key observation across all datasets is the expected reduction in accuracy following model optimization. Static quantization led to an average accuracy drop of 2.37\%, while dynamic quantization resulted in a slightly higher drop of 3.14\%. Similarly, L1 pruning and L2 pruning introduced accuracy degradations of 3.43\% and 3.82\%, respectively. However, these reductions in predictive performance were offset by substantial gains in inference speed and energy efficiency, suggesting that the trade-offs are acceptable for applications where computational efficiency is a priority.

\subsection{Inference Time and Energy Efficiency}
The inference time improvements were particularly notable. Quantization techniques reduced inference latency by factors of 1.42× (static quantization) and 1.52× (dynamic quantization), while pruning methods achieved even greater speed-ups, with L1 pruning reaching 1.63× improvement over the baseline. The energy savings associated with these optimizations were also substantial, with dynamic quantization reducing energy consumption by up to 33.25\% and L1 pruning achieving the highest energy savings of 37.08\%. This demonstrates that while accuracy slightly degrades, the impact on computational efficiency is significant, making these approaches viable for deployment in resource-constrained environments.

\subsection{Energy-Accuracy Trade-off Analysis}
The energy-accuracy trade-off analysis further supports the effectiveness of these optimizations. While baseline models exhibit the highest accuracy, their energy efficiency is considerably lower than that of optimized configurations. For instance, in the T1 model configuration, static quantization improved energy efficiency to 0.150 GFLOPS/J, while L1 pruning further enhanced it to 0.168 GFLOPS/J. The T2 model, which generally outperforms T1 in accuracy, also demonstrated improved energy efficiency through quantization and pruning, achieving an overall efficiency score of 26.63 after L1 pruning. These findings indicate that a carefully selected combination of quantization and pruning can provide the best balance between efficiency and accuracy retention.

\section{Conclusion}

Our study systematically investigates energy-efficient optimization strategies for transformer-based time series classification. By integrating structured pruning and quantization, significant improvements are achieved in inference speed and energy consumption while preserving robust classification performance. Notably, the proposed framework adapts to varying dataset characteristics, demonstrating that even models with higher computational complexity can be effectively compressed for deployment in resource-constrained environments. Our results underscore the need to tailor optimization strategies to specific application domains and offer valuable insights for developing scalable, sustainable deep learning solutions in edge computing scenarios.

\bibliographystyle{plain} 
\bibliography{references}

\end{document}